\documentclass{llncs}
\usepackage{cancel,enumitem,multirow,float,wrapfig,caption,color,subcaption}
\usepackage{array,graphicx,makeidx,amssymb,amsmath,subfig,booktabs}
\usepackage[utf8]{inputenc}

\begin{document}
\frontmatter
\pagestyle{headings}
\mainmatter
\title{On When and How to use SAT\\to Mine Frequent Itemsets}
\titlerunning{Using SAT to Mine Frequent Itemsets}
\author{Rui Henriques \and Inês Lynce \and Vasco Manquinho\vskip -0.1cm
\institute{INESC-ID/IST, Technical University of Lisbon, Portugal\\\vskip 0.2cm
\email{rmch@ist.utl.pt,\{ines,vmm\}@sat.inesc-id.pt}}}
\maketitle

\begin{abstract}
A new stream of research was born in the last decade with the goal of mining itemsets of interest using Constraint Programming (CP). This has promoted a natural way to combine complex constraints in a highly flexible manner. Although CP state-of-the-art solutions formulate the task using Boolean variables, the few attempts to adopt propositional Satisfiability (SAT) provided an unsatisfactory performance. 
This work deepens the study on when and how to use SAT for the frequent itemset mining (FIM) problem by defining different encodings with multiple task-driven enumeration options and search strategies. Although for the majority of the scenarios SAT-based solutions appear to be non-competitive with CP peers, results show a variety of interesting cases where SAT encodings are the best option. 
\end{abstract}

\section{Introduction}

Recent works \cite{key1,nary,key2,basis} show the cross-fertilization between \emph{Pattern Mining} (PM) tasks, the discovery of patterns within large datasets, and \emph{Constraint Programming} (CP), the programming paradigm wherein relations between variables are stated declaratively in the form of constraints. 
Traditional greedy approaches for PM contrast with optimal approaches developed within the artificial intelligence community. While traditional research aims at developing highly optimized and scalable implementations that are tailored towards specific tasks, CP employs a generic and declarative approach to model and mine patterns. 
This has motivated the adoption of high-level modeling languages or general solvers (that specify \emph{what} the problem is, rather than outlining \emph{how} a solution should be computed) for the flexible definition of constraints, which are critical for many PM applications and domains \cite{key0,nary,kpm}.

The core underlying task for every PM task is to count. Counting is required for every constraint: a specific pattern shape is only of interest above a minimum support threshold. However, here resides the efficiency bottleneck of CP solvers, which need to deal with large counting options to solve frequency-based inequalities. Even though state-of-the-art CP-based solutions are not yet as scalable as traditional PM solutions \cite{key1}, they can be used for local scans \cite{dmsat}, for the expressive definition of user-driven and non-trivial constraints \cite{expressive}, and their optimal search nature have led to significant performance improvements in a wide-diversity of problems \cite{key0}. 

In this work, we propose to study whether a specific class of CP solvers is tailored to solve this core task. The target class of solvers, propositional Satisfiability (SAT) solvers, aim to solve the Boolean SAT problem, which is the problem of finding an assignment for a set of Boolean variables that evaluates a target formula (usually restricted to a conjunctive normal form) to true. Although Boolean encodings are commonly adopted by CP solvers for PM tasks \cite{key2}, to the best of our knowledge SAT-based solutions have only been proposed to a particular subclass of PM problems aiming at discovering a fixed number of patterns (\emph{k}-PM) \cite{expressive,kpm}. 

Based on the critical need for an efficient CP solver for PM tasks, this work undertakes an extensive review to understand how SAT solutions compare to state-of-the-art CP-based alternatives. The target research question is: how SAT behaves in comparison to more general CP frameworks for PM underlying tasks? In section 2 the problem is defined and motivated. Section 3 introduces different SAT encodings with multiple enumeration and search options. Section 4 describes the properties of the adopted implementations and conducts an experimental analysis. The results are discussed and their implications synthesized. Section 5 reviews related research with relevant contributions to the target problem. Finally, concluding remarks and potential prospective research directions are presented.

\section{Problem Definition}



\begin{definition}
Let $\mathcal{I}$ be a finite set, called the set of items, let $\mathcal{T}$ be a finite set, called the set of transactions, 
and let $I$ be an itemset, $I\subseteq \mathcal{I}$. A \textbf{transaction} $t\in \mathcal{T}$ over $\mathcal{I}$ is a pair $(t_{id},I)$, with $t_{id}$ an identifier and $I\subseteq \mathcal{I}$. 
\end{definition}

\begin{definition}
An \textbf{itemset database} $D$ over $\mathcal{I}$ is a finite set of transactions. 
In a simplified way, a transactional dataset is a multi-set of itemsets (being the language of itemsets $L_{\mathcal{I}}  = 2^{\mathcal{I}}\backslash \emptyset$). 
Equivalently, an itemset database $D$ can be seen as a binary matrix of size $m\times n$, where $m$=$\mid$$\mathcal{T}$$\mid$ and $n$=$\mid$$\mathcal{I}$$\mid$, with $D_{ti} \in \{0, 1\}$, such that:

\begin{equation}
D = \{(t, I) | t \in \mathcal{T}, I \subseteq \mathcal{I}, \exists _{i \in I} D_{ti} = 1\} 
\end{equation}
\end{definition}

\begin{center}
\begin{table}
\scriptsize
\begin{center}
\begin{tabular}{c c}\toprule
& \textbf{Itemset}\\ \midrule
t$_1$ & A,E,G,K,N\\ 
t$_2$ & C,E,H,L,N\\
t$_3$ & A,D,H,J,O\\
t$_4$ & B,D,H,J,N\\
t$_5$ & A,D,H,J,N,P\\ 
t$_6$ & A,E,G,K,N,P\\ \bottomrule
\end{tabular}
\quad
\begin{tabular}{ccccccccccccccccccc}\toprule
 & A & B & C & D & E & F & G & H & I & J & K & L & M & N & O & P\\ \midrule
t$_1$ & 1 & 0 & 0 & 0 & 1 & 0 & 1 & 0 & 0 & 0 & 1 & 0 & 0 & 1 & 0 & 0\\ 
t$_2$ & 0 & 0 & 1 & 0 & 1 & 0 & 0 & 1 & 0 & 0 & 0 & 1 & 0 & 1 & 0 & 0\\ 
t$_3$ & 1 & 0 & 0 & 1 & 0 & 0 & 0 & 1 & 0 & 1 & 0 & 0 & 0 & 0 & 1 & 0\\ 
t$_4$ & 0 & 1 & 0 & 1 & 0 & 0 & 0 & 1 & 0 & 1 & 0 & 0 & 0 & 1 & 0 & 0\\ 
t$_5$ & 1 & 0 & 0 & 1 & 0 & 0 & 0 & 1 & 0 & 1 & 0 & 0 & 0 & 1 & 0 & 1\\
t$_6$ & 1 & 0 & 0 & 0 & 1 & 0 & 1 & 0 & 0 & 0 & 1 & 0 & 0 & 1 & 0 & 1\\ \bottomrule
\end{tabular}
\caption{Illustrative itemset database: compact and Boolean views}
\end{center}
\end{table}
\end{center}
\normalsize 

A small example of an itemset database is given in Table 1. A traditional example of an itemset database is the supermarket shopping, where each transaction corresponds to a transacted basket and every item to a bought product. However, common attribute-value tables can be easily converted into an itemset database. Both categorical data (where every attribute-value pair corresponds to an item) and numeric data (following an expressive discretization technique \cite{dmsat}) can be converted, with each row being mapped into a transaction. 

\begin{definition}
The \textbf{coverage} $\varphi_{D(I)}$ of an itemset $I$ is the set of all transactions in which the itemset occurs:
\begin{equation}
\varphi_{D(I)} = \{t \in \mathcal{T}\mid \forall_{i \in I} D_{ti} = 1\}
\end{equation}
\end{definition}
\begin{definition}
The \textbf{support} of an itemset $I$, denoted $sup_D(I)$, is its coverage size: $\mid$$\varphi_{D(I)}$$\mid$, and the \textbf{frequency} of an itemset, denoted $freq_D(I)$, is ${}^{sup_D(I)}/_{n}$.\vskip 0.08cm
\end{definition}

Considering the itemset database from Table 1, we have $\varphi_D(\{J,N\}) = \{t_4, t_5\}$, $sup_D(\{J,N\}) = \mid$$\{t_4, t_5\}$$\mid = 2$ and $freq_D(\{J,N\}) = 0.3(3)$. Now the target PM problem can be formulated.

\begin{definition}
Given an itemset database $D$ and a minimum support threshold $\theta$, the \textbf{frequent itemset mining} problem consists of computing the set:
\begin{equation}
\{I \mid  I \subseteq \mathcal{I}, sup_D(I) \geq \theta\}
\end{equation}
\end{definition}

\begin{definition}
Let a frequent itemset be an itemset with $sup_D(I) \geq \theta$, a \textbf{pattern} is a frequent itemset that satisfies any other placed constraints over $D$.
\end{definition}

Considering the database from Table 1 and fixing $\theta$=3, $\{D,H,J\}$ and $\{E,N\}$ are examples of frequent itemsets. Finding frequent itemsets is the core underlying task of every pattern discovery problem and, additionally, form the basis for association-rule analysis, classification, regression, and clustering. FIM was initially proposed in 1993 by Agrawal et al. \cite{fim}. 


\subsection{CP Mapping}

Flexible constraint-based methods are key for PM as they:
\begin{itemize}
\item Focus on \emph{what} the problem is, rather than outlining \emph{how} a solution should be computed, is powerful enough to be used across a wide variety of applications and domains \cite{key0,nary,kpm} as it suppresses the need of adapting the underlying traditional procedures in order to accommodate new types of constraints.
\item Provide an easy method to adapt the search by changing the declarative specification to combine and add new constraints. This not only supports user-driven selection of which patterns are of interest, but FIM-based methods that may require iterative refinements as constraint-driven clustering and pattern-based classification.
\item Can expressively capture background knowledge to prune the explosion of spurious and potentially non-interesting patterns \cite{mine}. These constraints may include properties from both closed pattern mining \cite{constraintPM} and domain-driven pattern mining \cite{onto4AR0}, which aim to incrementally improve results by refining the way patterns and domain-knowledge is represented.
\item Support the introduction of a wide variety of expressive constraints, as pattern-set constraints (e.g. global patterns imposing overlapping relaxations over local patterns), with key implications on a wide-variety of problems ranging from web mining to bioinformatics \cite{basis,key2}.
\end{itemize}

Although CP models like ConQueSt \cite{constraintPM} or MusicDFS \cite{musicDFS} already support a predefined number of constraints, they do not allow for the expressive definition and combination of constraints as CP approaches like FIMCP \cite{key2}, PattCP \cite{kpm}, or GeMini \cite{expressive}. The selected mapping is based on the constraint encodings of the later approaches. In these CP models for the FIM task, a Boolean variable is used for every individual item $I_i$ and for every transaction $T_t$. One assignment of values to all $I_i$ and $T_t$ corresponds to one itemset and its corresponding transaction set.

\begin{definition}
Let $T$ be a transaction set, $T\subseteq \mathcal{T}$. An \textbf{itemset} $I$ can be defined by the true item variables: $I_i = 1$ if $i \in I$ and $I_i = 0$ if $i\notin I$. A \textbf{transaction set} $T$ can be defined by the set of transactions that are covered by the itemset, $T = \varphi_D(I)$. Thus, $T_t = 1$ if $t\in \varphi_D(I)$. 
\end{definition}

\begin{corollary}
The \textbf{FIM} task can now be viewed as the computation of valid and frequent $(I,T)$ combinations, i.e. on finding the set:
\begin{equation}
\{(I,T) \mid I \subseteq \mathcal{I}, T \subseteq \mathcal{T}, T = \varphi_D(I), \mid T \mid \geq \theta\}
\end{equation}
\end{corollary}

We refer to $T = \varphi_D(I)$ as the \textbf{coverage} condition while $\mid T \mid \geq \theta$ expresses a \textbf{support} condition. These conditions restrict the valid variable assignments. Note that given that neither $I$ nor $T$ are fixed, there can be an arbitrary high number of valid attributions to $I_i$ and $T_t$ resulting in different $(I,T)$ tuples that satisfy both constraints. 

\begin{property}
[\textbf{Coverage Constraint}] Given a database $D$, an itemset $I$ and a transaction set $T$, then
\begin{equation}
T = \varphi_D(I) \Leftrightarrow (\forall_{t \in \mathcal{T}} : T_t = 1 \leftrightarrow \Sigma_{i\in I} I_i (1-D_{ti}) = 0)
\end{equation}
where $I_i\in\{0,1\}$, $T_t\in\{0,1\}$ and $I_i = 1$ if $i\in I$ and $T_t = 1$ if $t\in T$ \cite{key1}. 
\end{property}

\begin{property}
[\textbf{Frequency Constraint}] Given a database $D$, a transaction set $T$ and a threshold $\theta$, then 
\begin{equation}
\mid T \mid \geq \theta \Leftrightarrow \Sigma _{t\in \mathcal{T}}\geq \theta
\end{equation}
where $T_t\in \{0, 1\}$ and $T_t = 1$ if $t\in T$ \cite{key1}.
\end{property}

We can now model the frequent itemset mining problem as a combination of the coverage constraint and the frequency constraint. To illustrate this, \cite{key2} provides an example of a potential implementation in Essence \cite{essence} (solver-independent modeling language):\vskip 0.25cm

\textsf{1: given Freq : int, TDB : matrix[int(1..NrT),int(1..NrI)] of int(0..1)}

\textsf{2: find \textbf{I} : matrix[int(1..NrI)] of bool, \textbf{T} : matrix[int(1..NrT)] of bool}

\textsf{3: such that forall t: int(1..NrT)}

\textsf{4:\hspace{6 mm} \textbf{T}[t] $\Leftrightarrow$ ((sum i: int(1..NrI).(1-TDB[t,i]) \textbf{I}[i])<=0)} \emph{(Coverage Constraint)}

\textsf{5:\hspace{6 mm} (sum t: int(1..NrT).\textbf{T}[t])$\geq$Freq} \emph{(Frequency Constraint)}

\section{Mapping}

Since current state-of-the-art CP formulations for FIM rely on Boolean variables, it is important to understand the impact of adopting approaches dedicated to solve Boolean formulae. In particular, efficient and scalable SAT solvers developed over the last decades have contributed to dramatic advances in the ability to solve problem instances involving thousands of variables and millions of clauses \cite{expressive}. Additionally, a potential mapping of previous constraints into a conjunctive normal formula contains many binary clauses, which can be handled in a very effective manner by SAT solvers. For these reasons, this work encodes FIM as a SAT formula. Enumeration, search, encoding alternatives, and tunning options are covered.  



\subsection{Core Encoding}


The previously introduced coverage and frequency constraints map the FIM problem into a high-level CP language. Next, an extended SAT encoding is proposed. SAT clauses and pseudo-Boolean constraints will be interchangeably adopted to facilitate their traceability.

\begin{corollary}
[\textbf{Coverage Encoding}] Given a database $D$, an itemset $I$ and a transaction set $T$, the SAT encoding for the coverage constraint is:
\begin{equation}
\wedge_{t\in \mathcal{T}} (\wedge_{i\in I \mid D_{ti}} (\neg T_t \vee \neg I_i) \wedge (T_t\vee (\vee_{i\in I \mid D_{ti}} I_i)))
\end{equation}
\end{corollary}

\noindent\textbf{\emph{Proof}}. This formula is derived from equation (5) by:
\begin{enumerate}
\item rewriting the coverage sum: \vskip 0.13cm

$\Sigma_{i\in I} I_i (1-D_{ti}) = 0$ \vskip 0.12cm

$\leftrightarrow \Sigma_{i\in I \mid D_{ti}} I_i = 0$ \vskip 0.12cm

$\leftrightarrow \wedge_{i\in I \mid D_{ti}} \neg I_i$ \vskip 0.13cm

\item decomposing the equivalence into CNF: \vskip 0.13cm

$T_t=1 \Leftrightarrow \wedge_{i\in I \mid D_{ti}} \neg I_i$ \vskip 0.12cm

$\leftrightarrow (\neg T_t \vee (\wedge_{i\in I \mid D_{ti}} \neg I_i)) \wedge (T_t\vee \neg (\wedge_{i\in I \mid D_{ti}} \neg I_i))$ \vskip 0.12cm

$\leftrightarrow \wedge_{i\in I \mid D_{ti}} (\neg T_t \vee \neg I_i) \wedge (T_t\vee (\vee_{i\in I \mid D_{ti}} I_i))$ \vskip 0.13cm

\item encoding the quantifier $\forall _{t\in \mathcal{T}}^m$ into $m$ sets of clauses: \vskip 0.13cm

$\forall _{t\in \mathcal{T}} (T_t=1 \Leftrightarrow \wedge_{i\in I \mid D_{ti}} \neg I_i)$\vskip 0.12cm

$\leftrightarrow \wedge_{t\in \mathcal{T}} (\wedge_{i\in I \mid D_{ti}} (\neg T_t \vee \neg I_i) \wedge (T_t\vee (\vee_{i\in I \mid D_{ti}} I_i)))$\vskip 0.2cm
\end{enumerate}

\noindent\textbf{\emph{Complexity.}} Considering $n$ the size of $\mathcal{I}$ and $m$ the number of transactions in $D$, we have the following properties: an upper bound of $m\times n$ binary clauses, and $m$ clauses with a maximum of $n+1$ literals.\vskip 0.15cm

This SAT formula guarantees the consistency of $T_t$ and $I_i$ attributions.\vskip 0.1cm

To encode the \emph{frequency constraints}, we need to extend the SAT notation to include \emph{pseudo-Boolean} (PB) \emph{constraints}, which are extensions of SAT clauses that support cardinality constraints and weighted literals. Additionally, we need to adapt equation (6) into a reified frequency constraint for a more focused search of space, as discussed in \cite{key0}. This model is equivalent to the original model. 

\begin{property} 
[\textbf{Reified Frequency Constraint}] Given a database $D$, a transaction set $T$ and a threshold $\theta$, then:
\begin{equation}
\mid T \mid \geq \theta \Leftrightarrow (\forall_{i\in \mathcal{I}}: I_i=1 \rightarrow \Sigma _{T_t}D_{ti} \geq \theta)
\end{equation}
\end{property}

\begin{corollary} 
[\textbf{Frequency Encoding}] Given a database $D$, a transaction set $T$ and a threshold $\theta$, then the PB encoding for the frequency constraint is:
\begin{equation}
\wedge _{i\in \mathcal{I}} (\theta \neg I_i + \Sigma_{t\in T \mid D_{ti}} T_t \geq \theta)
\end{equation}
\end{corollary}

\noindent\textbf{\emph{Proof.}} This formula is derived from equation (8) by:
\begin{enumerate}
\item rewriting the frequency inequality into a cardinality constraint: \vskip 0.13cm

$\Sigma _{T_t}D_{ti} \geq \theta$ \vskip 0.12cm

$\leftrightarrow \Sigma_{t\in T \mid D_{ti}} T_t \geq \theta$ \vskip 0.13cm

\item decomposing the implication: \vskip 0.13cm

$I_i=1 \Rightarrow \Sigma _{T_t}D_{ti} \geq \theta$\vskip 0.12cm

$\leftrightarrow I_i=1 \vee (\Sigma_{t\in T \mid D_{ti}} T_t \geq \theta)$\vskip 0.12cm

$\leftrightarrow \neg I_i \vee (\wedge _{t\in T \mid D_{ti}} T_t \geq \theta)$\vskip 0.13cm

\item mapping the previous formula into conjunctive normal form or, as below, into pseudo-Boolean constraints:\vskip 0.13cm

$\neg I_i \vee (\Sigma_{t\in T \mid D_{ti}} T_t \geq \theta)$\vskip 0.12cm

$\leftrightarrow \neg I_i \vee (\frac{1}{\theta}\Sigma_{t\in T \mid D_{ti}} T_t \geq 1)$\vskip 0.12cm

$\leftrightarrow \theta \neg I_i + \Sigma_{t\in T \mid D_{ti}} T_t \geq \theta$\vskip 0.12cm

Potential SAT encodings for the $\neg I_i \vee (\Sigma_{t\in T \mid D_{ti}} T_t \geq \theta)$ constraint include the use of sequential counters \cite{sinz05}, binary decision diagrams \cite{minisatp}, sorting networks, or cardinality networks \cite{Asin11}. Note, however, that some of these encodings are not polynomial. Since solving these constraints is the core task of our problem, we benefit from solvers oriented to solve them. Moreover, a possible encoding using cardinality constraints would suffer from an additional complexity, since we would need to translate: $\neg I_i \vee Card_{Constraint}$. For these reasons, a pseudo-Boolean representation is the natural choice; \vskip 0.12cm

\item translating the quantifier $\forall _{i\in \mathcal{I}}^n$ into $n$ sets of clauses: \vskip 0.13cm

$\forall _{i\in \mathcal{I}} (I=1 \Rightarrow \Sigma _{T_t}D_{ti} \geq \theta)$\vskip 0.12cm

$\leftrightarrow \wedge _{i\in \mathcal{I}} (\theta \neg I_i + \Sigma_{t\in T \mid D_{ti}} T_t \geq \theta)$\vskip 0.2cm
\end{enumerate}


\noindent\textbf{\emph{Complexity.}} The incremental complexity added by this formula is $n$ pseudo-Boolean constraints (of the form \textsf{$\geq$}) with a maximum of $m$ unweighted literals and one weighted literal. 

Concluding, equations (7) and (9) describe the resulting SAT encoding, which has: \emph{i)} $m+n$ variables; \emph{ii)} $\Theta(mn)$ clauses with $mn$ being binary clauses and $m$ clauses having $\Theta(n)$ variables; and \emph{iii)} $n$ pseudo-Boolean constraints with $\Theta(m)$ variables. 

\subsection{Enumeration Options}

Using SAT solvers on the previously defined encoding would output either one frequent itemset ($I_i$ literals assigned as true) or \textsf{unsat}, meaning that no itemset satisfies the constraints. Understandably, model enumeration needs to be present in order to solve the FIM problem, that is, to find the set of all frequent itemsets. 

Note that an alternative strategy to define all frequent itemsets at the encoding level would mean an exponential growth of the search space, potentially behaving similarly to a simple enumeration strategy. However, since we can easily opt to adopt more expressive ways of performing enumerations, with significant space cuttings on each iteration, we only target enumeration strategies.

For this purpose, we need to introduce two properties of frequency that allow for pruning substantial parts of the search space, and dual formulation for FIM. 

\begin{definition}
Let $I$ be a set of items. A transaction $(t_{id}, J)$ contains $I$, denoted $I \subseteq (t_{id}, J)$, if $I \subseteq J$. FIM approaches rely heavily on:
\begin{itemize}
\item \textbf{monotonicity} of frequency: if $I \subseteq  J$, then the frequency of $J$ is bounded from above by the frequency of $I$; 
\item \textbf{anti-monotonicity} of frequency: if $I \subseteq J$ and $I$ is not frequent, then $J$ is also not frequent.
\end{itemize}
\end{definition}

\begin{definition}
Given an itemset database $D$ and a minimum support threshold $\theta$, the \textbf{dual-FIM} task centered on non-frequent itemsets is to compute:
\begin{equation}
2^\mathcal{I}/\{I \mid  I \subseteq \mathcal{I}, sup_D(I) < \theta\}
\end{equation}
\end{definition}

Table 2 rely on these notions to define different enumeration options. Explicit and compact negations based on the (anti-)monotonic property are proposed as well as further directions relying on alternative frequency-based properties.

\subsection{Search Options}

Previous enumeration strategies may not result in significant improvements if SAT iterations do not put any guarantee on the granularity of the itemsets found. For instance, if the subsets negation strategy is adopted and if the solver tends to output finer itemsets, the adoption of this strategy will not be relevant. The same is valid for the supersets negation strategy. To describe these search options, a new concept needs to be introduced.

\begin{definition}
A \textbf{maximal frequent itemset} is a frequent itemset that also satisfies:
\begin{equation}
\forall _{I' \supset I} : \mid \varphi (I')\mid < \theta
\end{equation}
\end{definition}

All itemsets that are a superset of a maximal itemset are infrequent, while all itemsets that are subsets are frequent. Maximal frequent itemsets are the top border between itemsets that are frequent and not frequent.

In an itemset database where $ABCD$ is the only maximal frequent itemset, a SAT solution using subsets negation may either find all frequent itemsets within $1$ iteration or across $\Sigma _{i=1}^{n=4}C^{n}_{i}$=15 iterations. The following three search options guarantee an upper bound on the number of performed iterations. 

\begin{center}
\scriptsize
\begin{tabular}{m{1.3cm} m{9cm} m{0.15cm} m{1.9cm}}\toprule
Option & Description & & Example $\mathcal{I}$$=$$\{A,B,C,D\}$\\ \midrule
Simple &
\textbf{\emph{Observation.}} Adding a clause with the negation of the solution on every iteration hampers the search, as it may result in redundant searches with multiple assignments of transaction-based literals $T_t$ for the same itemset. 

\textbf{\emph{Method.}} The added clause must only include the negation of item-based literals $I_i$ (i.e. exclude all transaction-based literals $T_t$). 
& & $sup(\{A,C\})$$\geq$$\theta$ $(\neg A$$\vee$$B$$\vee$$\neg C$$\vee$$D)$ \\ \midrule
Subsets Negation &
\textbf{\emph{Observation 1.}} If an itemset is frequent, then its subsets are also frequent (monotonicity). 

\textbf{\emph{Method 1.}} Add the negation of the found itemset as well as the negation of its subsets, so the number of SAT iterations can be largely reduced.

\textbf{\emph{Observation 2.}} Negating every frequent itemset may result in an impractical growth of the number of clauses during the search. For a very simple itemset database with $mn=100$, the number of added clauses can reach a thousand (significantly higher than the initial number of clauses used to encode the problem).

\textbf{\emph{Method 2.}} Compact the set of clauses obtained within one iteration in only one clause. That is, in next run the solver must be able to select, at least, one item that is not included in the previously found frequent itemsets:\vskip 0.1cm

\hspace{2cm}$I\geq \theta \Rightarrow \vee _{i\mid \neg I_i} I_i $ 
& & $sup(\{A,C\})$$\geq$$\theta$ $(\neg A$$\vee$$B$$\vee$$\neg C$$\vee$$D)$ $\wedge(\neg A$$\vee$$B$$\vee$$C$$\vee$$D)$ $\wedge(A$$\vee$$B$$\vee$$\neg C$$\vee$$D)$\vskip 0.8cm $sup(\{A,C\})$$\geq$$\theta$ $(B\vee D)$\vskip 0.5cm\\ \midrule
Supersets Negation &
\textbf{\emph{Observations.}} For fixed $I_i$ literal attributions satisfying the coverage constraints, if no combination of $T_t$ satisfies the frequency constraints, new clauses can be directly learned corresponding to the supersets of the found non-frequent itemset. This property was found to be critical for a dual-FIM formulation.

\textbf{\emph{Methods.}} Adopt dual-FIM problem for medium-to-low frequency thresholds, and negate supersets negation in a similar fashion as subsets negation (items of non-frequent itemsets cannot jointly appear): \vskip 0.1cm

\hspace{2cm}$I< \theta \Rightarrow \vee _{i\mid I_i} \neg I_i$ \vskip 0.08cm

Note that the choice of when to adopt the FIM-dual should be dynamically made based on the dataset properties (mainly density, but also transactions-to-items ratio) and on the inputted frequency.
& & $sup(\{A,C\})$$<$$\theta$ $(\neg A$$\vee$$B$$\vee$$\neg C$$\vee$$D)$ $(\neg A$$\vee$$\neg B$$\vee$$\neg C$$\vee$$D)$ $(\neg A$$\vee$$B$$\vee$$\neg C$$\vee$$\neg D)$ $(\neg A$$\vee$$\neg B$$\vee$$\neg C$$\vee$$\neg D)$ \vskip 0.25cm $sup(\{A,C\})$$<$$\theta$ ($\neg B\vee\neg D$)\\ \midrule 
Others & 
\textbf{\emph{Pointer 1.}} Adopt advanced enumeration strategy centered on implicit methods as, for instance, cube representations \cite{cube}.

\textbf{\emph{Pointer 2.}} Use the monotonicity and anti-monoticity properties to affect the exploitation of the structure of conflicts within a SAT solver.


\textbf{\emph{Pointer 3.}} Exploit important relationships between itemset frequencies beside monotonicity. These properties should not only affect the iterative insertions, but can be included as constraints in the initial encoding. This may result in significant improvements as the previous strategies. For example, in the MAXMINER algorithm \cite{freq0}, relations of the following
form are exploited: $freq(\{a,b,c\}) = freq(\{a,b\}) + freq(\{a,c\}) - freq(\{a\})$. 

There are many more relations between the frequencies of itemsets. See \cite{freq1} for extensions based on the inclusion-exclusion principle. For a generalization to other measures besides frequency, see \cite{freq2}. & & \textbf{--} \\ \bottomrule
\end{tabular}
\captionof{table}{\small Enumeration Options}
\end{center}
\normalsize 

\subsubsection{Largest-to-Shortest Maximal (LSM) search\\}


This search option guarantees that the number of iterations equals the number of maximal frequent itemsets by mapping the previous \emph{decision} problem into an \emph{optimization} problem. Each iteration returns a maximal frequent itemset, starting with the longest maximal frequent itemset until reaching the shortest maximal frequent itemset. This is done by defining the following goal function:

\begin{equation}
min: \Sigma _i^n \neg I_i,
\end{equation}

\noindent and, understandably, by adopting one of the expressive enumeration strategies based on the (anti-)monotonicity property. Otherwise, the goal of finding maximal frequent itemsets would be no longer a valid effort.

\subsubsection{Constrained Monotonic Growing (CMG) search\\}

One of the undesirable consequences of the LSM search method is the need to fully exploit the search space within each iteration. That is, for an optimum value for the goal function, several (potentially maximal) frequent itemsets need to be incrementally compared. Additionally, most of the learned behavior needs to be restarted across iterations. This is particularly critical for itemset databases with either many maximal frequent itemsets (usually the case) or with multiple fine maximal frequent itemsets. 

In order to overcome the referred problems, we propose CMG, a more relaxed search option formulated over decision tasks that do not require the larger maximal frequent itemsets to be found early. The CMG is based on two types of searches: \emph{$\alpha$-search} and \emph{$\beta$-search}. An $\alpha$-search is a simple SAT iteration (outputting one frequent itemset). As a result, not only a clause expressively negating its subsets is added, but also the itemset itself:\vskip 0.12cm

\small $temporaryClause \leftarrow \vee _{i\mid I_i}\{I_i\}$ \normalsize\vskip 0.12cm

After an $\alpha$-search, a set of $\beta$-searches are performed with the goal of finding larger frequent itemsets until a maximal frequent itemset is found (with \textsf{unsat} being returned). When this happens, the clauses related to the previous itemset's items are removed, the found frequent itemset is expressively negated and a new $\alpha$-search is performed. This behavior is repeated until the $\alpha$-search returns \textsf{unsat}. An illustrative instantiation segment of CMG behavior for an itemset database with $\mathcal{I}=\{A,B,C,D,E\}$ is presented next:\vskip 0.15cm

\small
$\{A\} \dashv \alpha$-$search()$\vskip 0.02cm

$temporaryClause \leftarrow (B\vee C\vee D\vee E)$\vskip 0.02cm

$\{A,B,E\} \dashv \beta$-$search(\{A,?\})$\vskip 0.02cm

$temporaryClause \leftarrow (C\vee D)$\vskip 0.02cm

$unsat \dashv \beta$-$search(\{A,B,E,?\})$\vskip 0.02cm

$learnedClauses \leftarrow learnedClauses \wedge temporaryClause$\vskip 0.02cm

$temporaryClause \leftarrow \emptyset$\vskip 0.02cm

$\{C,E\} \dashv \alpha$-$search()$
\normalsize

\subsubsection{Length Decreasing (LD) search\\}

A third search option, LD search, benefits from a more focused search of space as it fixes the length of the itemsets to be found. To guarantee that only maximal frequent itemsets are selected, LD initially fixes this length to $n$ and iteratively decrements it. Alternative length settings are possible if a separate initial scanning to the itemset database guarantees upper and lower bound restrictions on the length of maximal frequent itemsets.

LD accomplishes this behavior by adding and removing equalities of the form $\Sigma_iI_i=k$, with $k\in\{1,..,n\}$. However, since only few solvers support the addition-removal of pseudo-Boolean constraints, a new set of variables $A=\{A_1,..,A_n\}$ is added into the following additional constraint:
\begin{equation}
\Sigma_i^nI_i + \Sigma_i^nA_i=k
\end{equation}
In the simplest mode, all $A_i$ variables are initially assigned to \textsf{false} until no maximal frequent itemset with length $n$ is found (\textsf{unsat} output). Incrementally each $A_i$ is reversed to \textsf{true}, so finer maximal frequent itemsets can be found. Understandably, either the subsets negation enumeration option or the encoding of equation (11) needs to be in place for an efficient search.

\subsection{Mapping Restrictions}

The need to test the proposed options against different pseudo-Boolean solvers may require adaptations over the initial encoding. For simplicity, this section only describes how to support the removal of clauses during enumeration. In appendix, additional adaptations to deal with non-negated variables are covered. 

\begin{center}
\scriptsize
\begin{tabular}{m{1.6cm} m{11cm}}\toprule
Encoding & Description\\ \midrule

\emph{Incremental Clauses} &
\textbf{\emph{Method.}} Recur to: \emph{i)} $n$ additional variables ($N=\{N_1,..,N_n\}$), to \emph{ii)} $n$ additional clauses:
\begin{equation}
\wedge _{i\in I} (I_i \vee N_i)
\end{equation}
and to \emph{iii)} manipulations over the vector of assumptions that SAT solvers usually disclose.

In $\alpha$-searches the $N$ Boolean vector of assumptions is set to \textsf{true}, so these new $n$ clauses are directly satisfied. 
In $\beta$-searches, the index $i$ of the $I_i$ items belonging to the target itemset are fixed, and the respective $N_i$ variables are set to \textsf{false}. Recurring to $N_i$ assumptions, the SAT solver behavior is similar to solvers that allow for clause removal and the level of performance is closely maintained.

This is referred as incremental clauses encoding, because although no clauses need to be deleted, still new clauses need to be added.\\ \midrule

\emph{Fixed Clauses} &
\textbf{\emph{Observation.}} As the number of inserted clauses grows significantly with the number of iterations, an encoding with an increasing number of clauses penalizes the performance. 

\textbf{\emph{Method.}} The strategy non-prone to insertions requires part of the reasoning to be done outside of the solver. 
The challenge is that all new clauses are relevant, and, thus, need to be maintained in memory. For instance, in a $n$=5 itemset database, if the solver finds in initial iterations $\{I_1I_2I_3\}$ and $\{I_1I_3I_5\}$ (i.e. learned clauses are, respectively, $I_4\vee I_5$ and $I_2\vee I_4$), the new solutions need to satisfy all the learned clauses. In this strategy, the storage and reasoning is done separately to affect the values of the $N$ Boolean vector. Following the introduced example, either $N_4$ or $N_2$ and $N_5$ will iteratively assume the value \textsf{false}. So the binary clauses to be satisfied, described in the previous option, will guarantee that the respective items will be \textsf{true}.

This method of defining assumptions, needs however to be complemented with a control variable $y$. This control variable is required for CMG search option to distinguish between the $\alpha$- and $\beta$-searches. Additionally, the following constraint needs to be satisfied: 
\begin{equation}
y + \Sigma I_i + \Sigma N_i \geq n+1,\end{equation}
to guarantee that in $\beta$-searches ($y$ assumed to be false) an additional item is selected.\\\bottomrule 
\end{tabular}
\captionof{table}{\small Encoding Options for Clause Removals}
\end{center}
\normalsize 

The majority of available pseudo-Boolean solvers do not support the dynamic insertion or removal of clauses. Since the insertion of clauses is critical, when a solver of interest does not allow for insertion, its implementation needs to be adapted (usually by turning visible invocations to SAT solver methods at the level of the pseudo-Boolean solver interface). Although the removal of clauses is required within CMG and LD search options, it is rarely allowed and even non-easily disclosed recurring to the interface of SAT solvers. Two encoding adaptations to deal with removals are depicted in Table 3.

\subsection{Tunning Options}

Although multiple options for an efficient FIM were introduced, additional improvements can be performed either by adapting the initial encoding or the solver behavior. 

\subsubsection{Constraints Reduction\\}
From the multiple encoding adaptations that were studied, only one resulted in a significant performance improvement (for medium-to-high support thresholds). This adaptation is centered on a FIM-property that can lead to the reduction of the initial $\Theta(mn)$ constraints or binary clauses (if the solver is able to clausify all of these constraints) into only $\Theta(m+n)$ constraints: \vskip 0.12cm

$\wedge_{t\in T} (\neg T_t \vee (\wedge_{i\in I \mid D_{ti}} \neg I_i))$ \vskip 0.1cm

$\leftrightarrow \wedge_{t\in T} (\wedge_{i\in I \mid D_{ti}} \xcancel{(\neg T_t \vee \neg I_i)})$\vskip 0.1cm

$\leftrightarrow \wedge_{t\in T} (\mid$$D_{ti}$$\mid \neg T_t + \Sigma_{i\in I \mid D_{ti}}$-$I_i \geq 0)$ 

\subsubsection{Polarity Suggestions and Parameters\\} 
A simple and effective way to adapt the solver behavior is to change the polarity suggestions. Since we are interested in the early finding of larger itemsets, polarity suggestions for $I_i$ variables should be set to positive (this only degrades performance when maximal frequent itemsets have a very fine length). Orthogonally, the polarity suggestions for $T_t$ variables depend on a wide variety of factors (as the given frequency, density, iteration step and search strategy), and, therefore, can be dynamically attributed in a scope-sensitive manner.

Additionally, solver parameters as the decay factors for variable activity and clause activity can be dynamically tuned on the basis of sensitive analysis. 

Finally, the solver resolution can be adapted to be, for instance, sensitive to the difference between the transaction and item literals in a way that promotes a more focused search. This can result in significant performance improvements. Note, however, that the solver functionally must be ensured to support the addition of new flexible constraints.

\section{Results}

This section details the undertaken evaluation of the previous options against state-of-the-art CP solutions. First, we visit the properties of our implementation, then we describe the most significant observations, and, finally, we discuss the results to retrieve a set of implications.

\subsubsection{Covered Options\\} 
The supported encodings are the target FIM encoding and its dual formulation. The simple and expressive subsets and supersets negation are covered enumeration options. The modeling of (anti-)monotonicity properties at the encoding level are not supported since they imply an exponential growth on the number of variables and constraints. 
All the advanced search options (simple, LSM, CMG and LD) are supported as well. Additionally, all the restrictions introduced were found when linking some solvers, and, therefore, they are addressed in our experiments. Finally, an extensive set of tunning options were implemented, with the most significant being the ones described in the previous section.

\subsubsection{Codification Alternatives\\}
Different codifications were defined with the goal of supporting an efficient and flexible interface with multiple pseudo-Boolean solvers. For instance, a codification in Java can only interface efficiently with solvers in Java through direct invocation. Otherwise, interaction needs to be done between executables (the developed layer and the solver), leading to an additional latency as a result of the required synchronization between them. Additionally, the exchanged information requires parsing. This hampers the performance as information is extensively exchanged in every iteration (note that the number of iterations is usually greater than $n^2$ for low frequencies).

Two classes of SAT solvers were adopted. The first class comprises the solvers with open-source for whom all the covered options were implemented. The second class includes the solvers with undisclosed-source with whom a simple goal was assessed: see how they performed for specific single-iterations against the alternatives. This was justified by the fact that since most of them do not support neither the use of assumptions nor the insertion of new clauses required to perform enumerations. Although the full-feeding of these solvers with the encoding for every iteration was tried, the fact that they do not keep the learned clauses in memory turned their performance impractical. These solvers are PBS \cite{pbs}, BSOLO \cite{bsolo} and WBO \cite{wbo}. 

The adopted solvers belonging to the first class are SAT4J \cite{sat4j} and MiniSat+ \cite{minisatp}. Note that, since SAT4J is implemented in Java and MiniSat+ is implemented in C++, two codifications for the target solution were supported: under Java and C++\footnote{web.ist.utl.pt/rmch/research/software\\An additional third codification is available in C\# upon request.}.

\subsubsection{Datasets\\}
The adopted datasets were taken from the UCI repository\footnote{http://archive.ics.uci.edu/ml/}. The density of the dataset is defined by the average number of items per transaction divided by the size of the items' alphabet. Although the selected datasets are not large (note that optimal approaches suffer from scalability problems), they are dense by nature and, therefore, their use within traditional approaches is still largely computationally expensive.

\subsection{Observations}

The computer used to run the experiments was an Intel Core i5 2.80GHz with 6GB of RAM. The algorithms were implemented using Java (JVM version 1.6.0-24) and C++ (GCC 4.4.5) in 64-bit Linux (Debian 2.30.2) operating system.

\subsubsection{Comparative analysis\\}

Table 4 synthesizes the main results over UCI datasets for tunned MiniSat+ and SAT4J implementations under CMG search option and for the state-of-the-art CP performer, FIMCP. The proposed SAT-based solutions have a phase transition that relaxes for very low and medium-to-high frequency thresholds, as illustrated in Fig.1. Contrasting, FIMCP behavior increasingly deteriorates with the decrease of the frequency threshold. 

\begin{center}
\scriptsize
\begin{tabular}{lrrrrrrrrr}
\toprule
    {\bf } &            &            &            &          \multicolumn{ 3}{c}{$\theta$=0,02} &          \multicolumn{ 3}{c}{$\theta$=0,05} \\ 
\midrule
{\bf Dataset} &   \hspace{1 mm}  $\sharp$Items &   $\sharp$Trans &  Density &  \hspace{4 mm} MiniSat+ &      SAT4J &      FIMCP &  \hspace{3 mm} MiniSat+ &      SAT4J &      FIMCP \\
\midrule
{\it Tic-tac-toe} &         27 &        958 &       0,33 &     366,7 &    \textbf{--}$^\star$ &      0,2 &     550,0 &    \textbf{--}$^\star$ &       0,1 \\
{\it Primary-tumor} &         31 &        336 &       0,48 &      17,8 &    \textbf{--}$^\star$ &     1,0 &     300,0 &    \textbf{--}$^\star$ &      0,5 \\
{\it Zoo-1} &         36 &        101 &       0,44 &        0,3 &       4,3 &     1,5 &        0,8 &       7,2 &        0,6 \\
{\it Vote} &         48 &        435 &       0,33 &     528,5 &    \textbf{--}$^\star$ &     2,0 &    \textbf{--}$^\star$ &    \textbf{--}$^\star$ &      0,7 \\
{\it Soybean} &         50 &        630 &       0,32 &    \textbf{--}$^\star$ &    \textbf{--}$^\star$ &     1,9 &    \textbf{--}$^\star$ &    \textbf{--}$^\star$ &      0,5 \\
\midrule
{\it Hepatitis} &         68 &        137 &        0,5 &      60,5 &    \textbf{--}$^\star$ &    \textbf{--}$^\star$ &     229,8 &    \textbf{--}$^\star$ &    \textbf{--}$^\star$ \\
{\it Lymph} &         68 &        148 &        0,4 &    \textbf{--}$^\star$ &    \textbf{--}$^\star$ &    \textbf{--}$^\star$ &    \textbf{--}$^\star$ &    \textbf{--}$^\star$ &    \textbf{--}$^\star$ \\
{\it Kr-vs-kp} &         73 &       3169 &       0,49 &     380,5 &    \textbf{--}$^\star$ &    \textbf{--}$^\star$ &    \textbf{--}$^\star$ &    \textbf{--}$^\star$ &    \textbf{--}$^\star$ \\
{\it Hypothyroid} &         88 &       3247 &       0,49 &     433,3 &    \textbf{--}$^\star$ &    \textbf{--}$^\star$ &    \textbf{--}$^\star$ &    \textbf{--}$^\star$ &    \textbf{--}$^\star$ \\
{\it Heart-cleveland} &         95 &        296 &       0,47 &    \textbf{--}$^\star$ &    \textbf{--}$^\star$ &    \textbf{--}$^\star$ &    \textbf{--}$^\star$ &    \textbf{--}$^\star$ &    \textbf{--}$^\star$ \\
\midrule
{\it German-credit} &        112 &       1000 &       0,34 &    \textbf{--}$^\star$ &    \textbf{--}$^\star$ &    \textbf{--}$^\star$ &    \textbf{--}$^\star$ &    \textbf{--}$^\star$ &    \textbf{--}$^\star$ \\
{\it Mushroom} &        119 &       8124 &       0,18 &    \textbf{--}$^\star$ &    \textbf{--}$^\star$ &    \textbf{--}$^\star$ &    \textbf{--}$^\star$ &    \textbf{--}$^\star$ &    49126 \\
{\it Aust-credit} &        125 &        653 &       0,41 &    \textbf{--}$^\star$ &    \textbf{--}$^\star$ &    \textbf{--}$^\star$ &    \textbf{--}$^\star$ &    \textbf{--}$^\star$ &    \textbf{--}$^\star$ \\
{\it Audiology} &        148 &        216 &       0,45 &    \textbf{--}$^\star$ &    \textbf{--}$^\star$ &    \textbf{--}$^\star$ &    \textbf{--}$^\star$ &    \textbf{--}$^\star$ &    \textbf{--}$^\star$ \\
\bottomrule
\end{tabular}  
\vskip 0.5cm
\begin{tabular}{lrrrrrrrrr}
\toprule
    {\bf } &           \multicolumn{ 3}{c}{$\theta$=0,1} &           \multicolumn{ 3}{c}{$\theta$=0,2} &           \multicolumn{ 3}{c}{$\theta$=0,4} \\
\midrule
{\bf Dataset} & MiniSat+ &      SAT4J &      FIMCP &   \hspace{3 mm} MiniSat+ &      SAT4J &      FIMCP &  \hspace{3 mm} MiniSat+ &      SAT4J &      FIMCP \\
\midrule
{\it Tic-tac-toe} &      33,8 &      36,0 &       0,1 &       2,3 &     1,9 &     0,1 &      0,2 &       0,0 &       0,0 \\
{\it Primary-tumor} &   330,0 &    \textbf{--}$^\star$ &      0,2 &     126,2 &     124,0 &       0,1 &      38,0 &     9,5 &      0,0 \\
{\it Zoo-1} &       1,2 &       9,0 &      0,2 &     1,4 &    17,0 &      0,0 &      0,8 &     0,8 &       0,0 \\
{\it Vote} &    \textbf{--}$^\star$ &    \textbf{--}$^\star$ &      0,3 &     149,3 & 397,5 &      0,1 &     5,4 &    2,1 &      0,0 \\
{\it Soybean} &    \textbf{--}$^\star$ &     166,9 &      0,2 &     554,0 &    15,7 &       0,0 &      12,9 &     2,0 &      0,0 \\
\midrule
{\it Hepatitis} &    \textbf{--}$^\star$ &    \textbf{--}$^\star$ &    \textbf{--}$^\star$ &    \textbf{--}$^\star$ &    \textbf{--}$^\star$ &    2,5 &     382,1 &      \textbf{--}$^\dagger$ &      0,2 \\
{\it Lymph} &    \textbf{--}$^\star$ &    \textbf{--}$^\star$ &    19447 &    \textbf{--}$^\star$ &    \textbf{--}$^\star$ &      0,1 &      14,2 &     9,1 &       0,0 \\ 
{\it Kr-vs-kp} &    \textbf{--}$^\star$ &    \textbf{--}$^\star$ &    \textbf{--}$^\star$ &    \textbf{--}$^\star$ &    \textbf{--}$^\star$ &    \textbf{--}$^\star$ &    \textbf{--}$^\star$ &      \textbf{--}$^\dagger$ &    \textbf{--}$^\star$ \\
{\it Hypothyroid} &    \textbf{--}$^\star$ &    \textbf{--}$^\star$ &    \textbf{--}$^\star$ &    \textbf{--}$^\star$ &    \textbf{--}$^\star$ &    \textbf{--}$^\star$ &    \textbf{--}$^\star$ &    \textbf{--}$^\star$ &    \textbf{--}$^\star$ \\
{\it Heart-cleveland} &    \textbf{--}$^\star$ &    \textbf{--}$^\star$ &    \textbf{--}$^\star$ &    \textbf{--}$^\star$ &    \textbf{--}$^\star$ &    \textbf{--}$^\star$ &     508,5 &      \textbf{--}$^\dagger$ &      0,8 \\
\midrule
{\it German-credit} &    \textbf{--}$^\star$ &    \textbf{--}$^\star$ &    \textbf{--}$^\star$ &    \textbf{--}$^\star$ &    \textbf{--}$^\star$ &    15,9 &    \textbf{--}$^\star$ &    \textbf{--}$^\star$ &     0,5 \\
{\it Mushroom} &    \textbf{--}$^\star$ &    \textbf{--}$^\star$ &    11,9 &    \textbf{--}$^\star$ &    \textbf{--}$^\star$ &    2,9 &    99,9 &      \textbf{--}$^\dagger$ &     0,5 \\
{\it Aust-credit} &    \textbf{--}$^\star$ &    \textbf{--}$^\star$ &    \textbf{--}$^\star$ &    \textbf{--}$^\star$ &    \textbf{--}$^\star$ &    \textbf{--}$^\star$ &    \textbf{--}$^\star$ &    \textbf{--}$^\star$ &    5,5 \\
{\it Audiology} &    \textbf{--}$^\star$ &    \textbf{--}$^\star$ &    \textbf{--}$^\star$ &    \textbf{--}$^\star$ &    \textbf{--}$^\star$ &    \textbf{--}$^\star$ &    \textbf{--}$^\star$ &    \textbf{--}$^\star$ &    \textbf{--}$^\star$ \\
\bottomrule
\end{tabular}  
\captionof{table}{\small{Overall efficiency of the proposed solvers against FIMCP (seconds)\\ $^\star$timeout; $^\dagger$memory out;}}
\end{center}  
\normalsize

\begin{figure}[!htb]
\begin{subfigure}{0.58\textwidth}
\centering
    \includegraphics[width=1\textwidth]{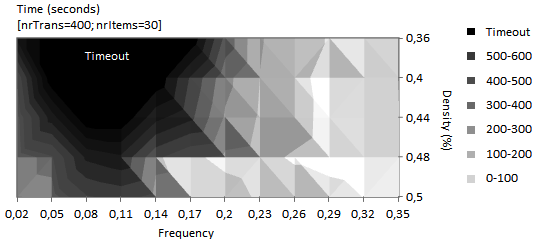}
\end{subfigure}
\begin{subfigure}{0.39\textwidth}
\centering
    \includegraphics[width=1\textwidth]{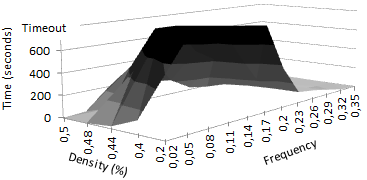}
\end{subfigure}
  \vspace{-5pt}
\caption{\small{Phase transitions for varying densities of generated datasets (using MiniSat+)}}
  \vspace{-5pt}
\end{figure}

FIMCP is the best option when targeting either medium-to-high frequency thresholds or normal-to-low dense datasets. However, when the target problem is based on low frequency thresholds over dense or large datasets, our adapted MiniSat+ solution is the choice. Note that these cases are the most common scenario in FIM problems, where the required frequency range to perform association rules falls between 1 and 2\%. An illustrative example of this advantageous behavior can be observed over the \textsf{hepatitis} dataset. The performance of FIMCP for this dataset is only scalable until frequency thresholds near and above 20\%. For $\theta<$10\%, FIMCP performance exponentially deteriorates with $\theta$ decrements. Interestingly, under this same $\theta$ range, MiniSat+ is able to answer to the FIM problem in useful time as shown in Fig.2.

\begin{figure}[!htb]
\begin{center}
    \includegraphics[width=0.62\textwidth]{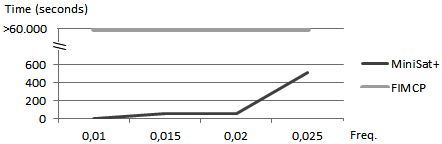}
  \end{center}
  \vspace{-7pt}
  \caption{\small{MiniSat+ behavior for \textsf{Hepatitis} dataset under very low frequencies}}
  \vspace{-5pt}
\end{figure}

Since the performance of SAT4J is hampered by its resolution properties (not tunned to deal with low frequency thresholds and not able to clausify key constraints) and by a bad memory management (dependent on a garbage collector), the adoption of MiniSat+ or FIMCP is overall preferred.

\subsubsection{Selecting SAT-based solvers\\}

The selection of best performer selections based on the inputted frequency and dataset properties implies an extensive analysis of the behavior of the solvers across different axes of choice. Table 5 synthesizes the main results of the undertaken analysis, which are further detailed in Appendix C.

\subsection{Deepening the analysis behind SAT vs. CSP}

Up to now, the most efficient CP approaches map FIM as a Constraint Satisfaction Problem (CSP) \cite{key1}. A CSP problem is specified by a finite set of variables $V$, an initial domain $D$ (which maps every variable $v\in V$ to a finite set of values $D(v)$), and a finite set of constraints $C$ in first-order logic. The goal is to output the variable domains which satisfy all constraints. Thus, the solution to a FIM problem can be directly retrieved from the codification of equations (5) and (8). 

\begin{center}
\scriptsize
\begin{tabular}{m{2.3cm} m{10cm}}
\toprule
Axis & Observations \\ \midrule
Enumeration \hspace{1cm}options & 
$\circ$ Although the negation of subsets or supersets within iterations lead to significant performance improvements, the level of impact depends on the search ability to early discovery the largest itemsets (as in LSM, CMG and LD); \vskip 0.1cm

$\circ$ Interestingly, the explicit negation of each subset/superset is preferred over the expressive one-clause-only negations when there is a high number of iterations as solvers are able to remove duplicated negated subsets, while in the later option there is an increasing redundancy that may hamper the search performance.
\\ \midrule
Search options & 
$\circ$ CMG search is overall the best choice ($\beta$-searches are efficient);\vskip 0.1cm

$\circ$ Simple search methods are only able to perform searches with acceptable efficiency on high frequency thresholds over very sparse datasets, and should only be adopted when resolution promotes an early discovery of larger itemsets;\vskip 0.1cm

$\circ$ LSM searches are only competitive when a few number of iterations is performed since each search is a lot heavier than a full CMG search (including one $\alpha$-search and multiple $\beta$-searches);\vskip 0.1cm

$\circ$ LD is competitive with CMG for medium frequency thresholds ($0.05<\theta <0.2$) on small datasets. LD performance quickly degrades with their growing size. Although \textsf{sat} searches are light (focused on discovering frequent itemsets with a given length) a significant overhead is added by the fixed number of heavier \textsf{unsat} searches, particularly for a low-to-medium number of maximal frequent itemsets.
\\ \midrule
Encoding options (restrictions) & 
$\circ$ Clause-oriented encodings seems to be preferred over minimal encodings for lower-to-medium frequency thresholds. Although minimal encodings have significantly fewer constraints, they are not easy to clausify; \vskip 0.1cm

$\circ$ Restricted encodings penalize the performance significantly (5-25\%).
\\ \midrule
Tunning options &  
$\circ$ Positive suggestions are adequate for low-to-medium frequencies (indicate a preference towards larger itemsets), while negative suggestions are indicated for higher frequencies (since large itemsets are not frequent, more conflicts are found increasing the number of backtracks and potential restarts);
\\ \midrule
Implementation options &
$\circ$ SAT4J solver is the option for high frequency thresholds and for very sparse datasets since it accepts non-constrained encodings and its resolution is more tuned to find finer itemsets;\vskip 0.1cm

$\circ$ MiniSat+ solver is the natural choice for the rest of the options due to, among others, the tunned parameterizations as polarity suggestions and decaying factors.\\ \bottomrule
\end{tabular}  
\captionof{table}{\small{Key observations for the target SAT-based solvers across six dimensions}}
\end{center}  


The first key concept used to speed up the search is constraint propagation to reduce the domains of variables such that the domain remains locally consistent. To maintain local consistencies, propagators are used to remove values from a domain that can never satisfy a constraint. 

Besides this property, constraint-based solvers are well-prepared to deal with certain types of constraints. Two examples are: the summation constraint and the reified summation constraint \cite{key0}. Flexible solvers as Gecode are well-prepared to deal with them. The FIM problem heavily relies on these constraints.

Let $x \in V \subseteq \mathcal{V}$ be a variable with an associated weight $w_x$, a summation constraint as the following form \cite{key0}:
\begin{equation}
\Sigma _{x\in V} w_x x \geq \theta\hspace{0.2cm}
\end{equation}
The propagator task is to discover as early as possible whether the constraint is violated (i.e. whether the upper-bound of the sum is still above the threshold). 

In a reified summation constraint, the evaluation of a summation constraint depends on a Boolean variable $b$ (as adopted in the target frequency constraints):
\begin{equation}
b \rightarrow C' (usually\hspace{0.2cm} C'=\Sigma _{x\in V} w_x x \geq \theta)
\end{equation}
The most important propagation that occurs for this constraint is the one that updates the domain of $b$ \cite{key0}. 
When an item variable is fixed, the following is possible for the coverage constraint:\vskip 0.15cm

if for some t: $\Sigma_{i\in I}(1 - D_{ti})I_i^{min} > 0$ then remove 1 from $D(T_t)$

if for some t: $\Sigma_{i\in I}(1 - D_{ti})I_i^{max} = 0$ then remove 0 from $D(T_t)$\vskip 0.1cm

Once the domain of a variable $T_t$ changes, the support constraint is activated. The support constraint is simply a summation constraint, which checks whether: $\Sigma _{t\in A}T_t^{max}\geq \theta$. If this constraint fails, CSP solvers do not need to branch further and, therefore, can backtrack. 

Contrasting, the mapping of reified frequency constraint into SAT is not concisely handled, leading to the generation of $nm$ clauses. Jointly these observations and the fact that SAT solvers are not able to expressively deal with enumerations (in particular, when an arbitrary number of clauses are added between iterations) justify the poor performance of the developed solutions across several datasets and frequencies.

\subsection{Discussion}
From the previous observations, several implications with impact on when to use and how to tune SAT-based solutions can be retrieved.\vskip 0.25cm

\textbf{\emph{When}} to use SAT-based solutions:
\begin{itemize}
\item for low frequency thresholds, with most promising results on dense datasets (for instance, datasets typically adopted for classification tasks with nominal attributes with few labels, or with numeric attributes that are binarized using thresholds). The level of frequency used to opt for a SAT-based solution depends on the density: if density is near 45-50\% the frequency can reach 10\%, while for other cases the frequency should not exceed 4\%;
\item when the problem is not defined as a complete enumeration, but aims to find a fixed number of patterns of interest, or to verify its satisfiability;
\item in specific cases for higher frequencies (mainly between 10\% and 25\%) when the FIM problem is formulated as its dual;
\end{itemize}

\textbf{\emph{How}} to tune SAT-based solutions:
\begin{itemize}
\item use an expressive enumeration strategy such as the compact subsets negation, change polarity suggestions (item-related variables to positive and transaction-related variables to false) and prefer clause-oriented encodings;
\item adopt the search strategy according to the target instances and problem:
\begin{itemize}
\item by default and for the majority of cases, CMG search is the most efficient; 
\item \emph{simple} search should only be adopted when the polarity suggestions can be set according to the proposed guidelines for datasets with a good distribution of frequent itemsets among transactions (otherwise the use polarity suggestions do not guarantee that, within each iteration, large itemsets are discovered);
\item LSM search should be selected when we are interested in a subset of itemsets with major interest (i.e. when there is the requirement of obtaining the $k$ largest maximal frequent itemsets). In particular, the value of $k$ should be significantly lower than the total number of maximal frequent itemsets (phase transition), otherwise a CMG search should be preferred (although there is the need to generate all maximal frequent itemsets since it does not guarantee if a maximal frequent itemset is one of the $k$ largest); 
\end{itemize}
\item be able to translate the flexible user-defined constraints into a SAT formula, by defining generic methods to translate equivalences and implications. Note that the distinguishing feature of CP is that it provides general principles for solving problems with any type of constraints. Although some streams claim that this observation sets it apart from SAT solving \cite{basis}, we show that this is a simple step as it is illustrated by the following constraints' encodings:
\begin{itemize}
\item $I^p = I^q \rightarrow  \wedge _{i\in\mathcal{I}}(I_i^p \Leftrightarrow  I_i^q)$
\item $i \in I^p \rightarrow  I_i^p$
\item $I^p \ I^q = I^r \rightarrow  \wedge _{i\in\mathcal{I}}(I_i^r\Leftrightarrow  I_i^p \wedge \neg I_i^q)$
\item $I^p \cap  I^q = I^r \rightarrow  \wedge _{i\in\mathcal{I}}(I_i^r \Leftrightarrow  I_i^p \wedge  I_i^q)$
\item $I^p \cup I^q = I^r \rightarrow  \wedge _{i\in\mathcal{I}}(I_i^r \Leftrightarrow  I_i^p \vee I_i^q)$
\item $coverItems(I^1, ..,I^k) \rightarrow \wedge _{i\in\mathcal{I}}(\vee _{j\in 1..k} I_i^j)$
\item $coverTrans(I^1, ..,I^k) \rightarrow \wedge _{t\in\mathcal{T}}(\vee _{j\in 1..k} T_t^j)$
\end{itemize}
\end{itemize}
\vskip 0.2cm

Unfortunately, both the proposed SAT-based solution and any other CP solutions are featured by high computational complexity and their straightforward implementations are not applicable to large data sets. 
This advocates the need for local learning where transactions are partitioned and multiple criteria can be used for the integration of the frequent patterns found within each fragment (for instance, through voting techniques \cite{dmsat}). Alternatively, the FIM constraints can be used to compute information gain metrics \cite{freqsat1} as the entropy-measure or its dual formulation as a basis for pruning techniques.

\section{Related research}

The main research streams approaching PM within a CP framework can be classified according to: \emph{i)} the extent to which an user can define novel constraints and combine them, and \emph{ii)} according to the type of supported constraints. Within the first axis, although approaches as Patternist \cite{constraintPM}, Molfea \cite{molfea} and MusicDFS \cite{musicDFS} support a predefined number of constraints, they do not allow for the expressive definition of novel constraints as FIMCP \cite{key2}, PattCP \cite{kpm} and GeMini \cite{expressive}. In the following section the constraints covered by existing approaches and variations to the FIM problem are briefly presented. Finally, other related work with potential relevant contributions is covered.

\subsubsection{Extending Constraints.\\}
Expressive CP models \cite{expressive,nary,kpm} enable the flexbile definition of constraints using: \emph{constants} including numerical values, items as $A$, specific itemsets as $\{A,B\}$, and specific transactions as $t_7$; \emph{variables} ($\vee_{i\mid C} I_i$ and $\vee_{t\mid C} T_t$); 
set and numerical \emph{operators} ($\cup, \cap, \setminus, \times, +, -$); and \emph{function symbols} involving one or several terms, which can be built-in as $overlapTrans(I,J)$$=$$\mid$$cover(I)\cap cover(J)$$\mid$) or defined by the user as $area(I)$$=$$freq(I)$$\times$$size(I)$ or $coverage(I,J)$=$freq(I$$\cup$$J)$$\times$$size(I$$\cap$$J)$.

These constraints can be used to filter the patterns of interest. However they can be used with a different purpose: closed itemset mining, discriminative pattern mining, pattern-based clustering and pattern-based classification. A recent direction is taking into account the relationships between local patterns to produce global patterns or pattern sets \cite{nary}. Despite their importance, there are very few attempts to mine patterns involving several local patterns and the existing methods tackle just particular cases by using devoted techniques \cite{key0}. In \cite{expressive,basis}, the importance of adopting a declarative CP-based approach to mine global patterns was highlighted by several examples coming from clustering tasks based on associations. Due to the complexity of this task and its easy modeling through constraints, CP-based approaches as pattern teams \cite{teams} have been providing encouraging results when compared against heuristic-based approaches that consider the added value of a new global pattern given extensive combinations of selected patterns \cite{heuristick}.

\subsubsection{Problem Variations.\\}
The introduced FIM approach can be extended and adapted in different ways. One of the most common is the $k$-pattern set mining \cite{kpm,expressive}. Unlike FIM, the problem of $k$-pattern set mining is concerned with finding a set of $k$ related patterns under constraints. The discovery of $k$ representative patterns often uses probabilistic models for summarizing frequent patterns \cite{condensed1} and other condensed representations of patterns as the dataset compression using Minimum Description Length Principle \cite{condensed2}. These approaches mainly aim at reducing the redundancy between patterns and, like our SAT approach, often focus on maximal frequent patterns. The $k$-pattern set mining problem is a very general problem that can be instantiated to a wide variety of mining tasks including concept-learning, rule-learning, re-description mining, conceptual clustering and tiling \cite{kpm,expressive}.



\subsubsection{Other Relevant Work.\\}
Instead of mining rules that rely on frequent itemsets, some approaches encode the problem within the CP framework using (compact) reducts \cite{classif}, i.e., subsets of most informative attributes. In this stream, a SAT representation is often formulated as an Integer Programming (IP) model to solve the minimal reduct constraints. In \cite{dmsat}, an extensive research is done over Boolean reasoning methodologies for Rough Set theory. Two problems are encoded: the search for reducts and the search for decision rules which are building units of many rule-based classification methods.

Another important direction is the verification whether PM constraints are satisfiable. This is a one-iteration-only decision problem that, according to our previous analysis, can be handled using SAT solvers. An example of this task with multiple applications in privacy preserving data mining, condensed representations and the FIM problem, is the called FREQSAT problem \cite{freqsat2}: given some itemset frequency-interval pairs, does there exist a database such that for every pair the frequency of the itemset falls into the interval? That is, given a set of frequency constraints $C = \{freq (I_j) \in [l_j , u_j], j = 1...m\}$, verify if there exists a database $D$ over $\cup ^m_{j=1} I_j$ that satisfies $C$ \footnote{suppose that the following set C of frequency constraints $C$ is given: $\{freq (\{a, b\})$ $\in [3/4, 1]; freq (\{a, c\}) \in [3/4, 1], freq (\{b, c\}) \in [3/4, 1], freq (\{d, e\}) \in [3/4, 1],$ $freq (\{d, f\}) \in [1/2, 1], freq (\{e, f\}) \in [1/2, 1], freq (\{a, b, c, d, e, f\}) = 0\}$; $C$ is in \textsf{FREQSAT}, because it is satisfiable by the following database: $D = \{(1,\{a, b, c, d, e\}),$ $(2,\{a, b, c, d, e\}),$ $(3,\{a, b, c, d, e\}),$ $(4,\{a, b, c, d, f\}),$ $(5,\{a, b, c, e, f\}),$ $(6,\{a, b, d, e, f\}),$ $(7,\{a, c, d, e, f\}),$ $(8,\{b, c, d, e, f\})\}$}. The problem can be further extended to include arbitrary Boolean expressions over items and conditional frequency expressions in the form of association rules. Additionally, FREQSAT is equivalent to probabilistic satisfiability (pSAT) \cite{freqsat1}.


\section{Concluding Remarks}

The use of SAT within the CP framework to address constrained-pattern mining problems was demonstrated to be valid: not only the adopted variables are Boolean, but expressive constraints can be easily mapped into a Boolean formulae. Multiple search options, enumeration strategies, encoding alternatives and parameterizations were studied in order to improve its performance. Understandably, these adaptations aim to orient SAT reasoning to the main properties of frequent itemset mining, without loosing the ability to flexibly support novel constraints.


Experimental results show that SAT-based solutions are competitive with the state-of-the-art CSP solutions for an important range of frequencies (range commonly adopted to discover association rules or to perform classification tasks). The efficiency problems found for higher frequencies mainly result from the fact that SAT was not developed with the intent of perform enumerations under the addition of new (and potentially conflicting) clauses. Finally, a set of guidelines were introduced to understand when to use and how to tune SAT-based solutions.


\subsubsection{Future Work.}
In the next steps one should expect:
\begin{itemize}
\item the extension of this approach to evaluate its impact not only in terms of efficiency but also in terms of accuracy for FIM-based problems as the discriminative-FIM problem;
\item the exploitation of SAT limits and performance in comparison to CP approaches under an intensive use of constraints;
\item the development of hybrid approaches with clear rules to select and tune the best performer under certain conditions: \emph{i)} the selected frequency, \emph{ii)} the nature of the problem constraints and relaxations, and \emph{iii)} the dataset properties -- by order of relevance: the density, the number of items, the item-to-transaction ratio and the number of transactions;
\item the exploitation of potential improvements from the codification of the adopted enumeration and search strategies at the encoding level;
\item the adoption of more expressive constraints besides (anti-)monotonicity, including extensions based on the inclusion-exclusion principle \cite{freq1} and other frequency-based relations \cite{freq0,freq2};
\item the finding of patterns in continuous data (as, for instance, required in many bioinformatic datasets), which may require discretization techniques beyond support-vectors, rough set and Boolean reasoning theories \cite{dmsat};
\item the mining of frequent itemsets in structured data as sequences, trees and graphs. New formulations are required to represent these problems under CP, which may not be trivial to encode when recurring to a fixed number of features or variables;
\item the assessment of SAT approaches to perform constraint-based clustering and constraint-based classifier induction (not necessarily relying on frequent itemsets). In constraint-based clustering the challenge is to cluster examples when additional knowledge is available about these examples, for instance, prohibiting certain examples from being clustered together (so-called cannot-link constraints) \cite{expressive,key1}. Similarly, in constraint-based classifier induction, one may wish to find a decision tree that satisfies size and cost-constraints \cite{mintree}. In traditional data mining, the relationship between itemset mining and constraint-based decision tree learning was studied in \cite{dmtree}, however such relation was not yet exploited in a CP setting;
\item the adoption of SAT verification (as the previously introduced FREQSAT problem \cite{freqsat1}) using deduction rules to prune/constrain the search of frequent itemsets. The monotonicity rule is a very simple example of deduction. More advanced rules, as the partial frequency available for some itemsets, bound on the frequencies of itemsets yet to be counted. Examples of deduction rules to improve pruning and speed-up FIM approaches are given in \cite{final0,final1};
\item the study of potential techniques to turn SAT and other CP-based approaches scalable. Options may include the local scanning on dataset partitions \cite{dmsat} or the use of data stream mining approaches \cite{stream}. 
\end{itemize}


\bibliographystyle{splncs03}

\appendix
\small
\section{Positive Literals}

Some solvers only admit a simplified and constrained pseudo-Boolean notation as input. An example is the exclusion of negated literals, which may not be trivial to handle. The translation of \emph{coverage} constraints into constraints with non-negated literals is trivial:\vskip 0.2cm

$\wedge_{t\in T} (\wedge_{i\in I \mid D_{ti}} (\neg T_t \vee \neg I_i) \wedge (T_t\vee (\vee_{i\in I \mid D_{ti}} I_i)))$\vskip 0.15cm

$\leftrightarrow \wedge_{t\in T} (\wedge_{i\in I \mid D_{ti}} (T_t \vee I_i \leq 1) \wedge (T_t\vee (\vee_{i\in I \mid D_{ti}} I_i) \geq 1))$\vskip 0.2cm

The \emph{frequency} constraints were translated with the goal of maintaining the advantageous $\geq$ operator:\vskip 0.2cm

$\wedge _{i\in I} (\theta \neg I_i + \Sigma_{t\in T \mid D_{ti}} T_t \geq \theta)$\vskip 0.15cm

$\leftrightarrow \wedge _{i\in I} (-\theta \neg I_i + \Sigma_{t\in T \mid D_{ti}} T_t \geq 0)$\vskip 0.2cm

In this fashion, solvers as \textsf{Minisat+} \cite{minisatp} can be tested without significant overhead, as most of them internally are able to clausify both of the pseudo-Boolean constraints defined for the coverage restrictions.

\section{Results}

The following appendix sections detail the observations made in Table 3. The adopted datasets for these experiments are either: real UCI datasets (distributed in Fig.3 according to their properties) or generated datasets (with customized number of items, number of transactions and density) whose generation depend on a biasing parameter $\gamma$ for the emerging of patterns according to a distribution similar as real datasets. The small but representative \textsf{zoo} dataset is often used to compare options.

\begin{figure}[!htb]
  \vspace{-5pt}
\begin{center}
\includegraphics[width=6.8cm]{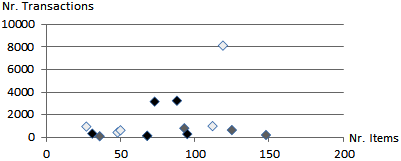}
\end{center}
\label{datasets}
  \vspace{-15pt}
\caption{\small{Characterization of the target datasets (high density in black, low in gray)}}
  \vspace{-15pt}
\end{figure}

\subsection{Enumeration options}

As expected, the adoption of the monotonicity principle within each enumeration have significantly impacted the performance of SAT solvers. According to an extract of the results in Table 6, two main observations can be drawn. \emph{First}, the level of impact by (either simply or expressively) negating subsets depends on the search option and on the target frequency. 

\begin{table}
\scriptsize
\begin{center}
\begin{tabular}{llrrr}\toprule
{\bf Search} & {\bf Enumeration} &        $\theta$=0,2 &        $\theta$=0,4 &      \hspace{2 mm}  $\theta$=0,6 \\ \midrule
\multirow{ 4}{*}{Optimization (SAT4J)\hspace{1 mm}} &     Simple &   \hspace{1 mm} 1514,0 &   \hspace{1 mm}  230,0 &  \hspace{2 mm}      1,0 \\ 
 & Expressive Simple &   943,7 &      89,2 &        0,1 \\ 
 & Subsets Negation &      31,3 &       2,6 &         0,1 \\ 
 & Expr. Subsets Neg. &      18,2 &       1,8 &        0,1 \\ \midrule
\multirow{ 4}{*}{Decision (SAT4J)} &     Simple &    1009,4 &      12,6 &        0,5 \\
 & Expressive Simple &     589,8 &       5,4 &        0,2 \\
 & Subsets Negation &     208,8 &       4,8 &        0,1 \\
 & Expr. Subsets Neg. &     183,0 &       6,5 &        0,2 \\ \bottomrule
\end{tabular}  
\caption{\small{Comparison of enumeration strategies for the \textsf{zoo} dataset (seconds)}}
\end{center}
\end{table}

Understandably, such impact depends on the ability to early discover the largest itemsets. This is the reason why subsets negation is key to LSM and CMG searches (its development relies on this principle), and important, but not so significant, within simple search methods. Additionally, cutting space through subsets negation is more critical for low frequencies, as an increasing length and number of itemsets is observed.

\emph{Second}, the choice of when to explicitly insert a negation of each subset or to expressively insert one unique clause requiring the selection of an item not observed in the found itemset is not simple. This choice depends on two factors: the total number of similar itemsets and the relative length of the found itemset. In the first case, when we have multiple similar frequent itemsets (sharing the majority of items), the explicit insertion of repeated negated sub-itemsets is detected by the solver, and the repeated clauses are removed, while in the expressive insertion every new clause is inserted as-is as a new problem constraint. In the second case, when the length of itemsets is small in comparison with the item alphabet this means that, although the negated sub-itemsets inserted by an explicit negation strategy generate multiple clauses, the number of clauses is not high and the number of literals per clause is low. Contrasting, although expressive negation strategies only add a clause per iteration, the number of literals can be significantly high and may hamper the resolution performance. These observations claim for an increased attention on the strategy selection based on the inputted frequency and dataset properties.

%

\subsection{Search options}

The first observation coming from the search option results is that, although CMG search is overall the best choice, the performance of the searches highly vary according to the dataset density and input frequency. Maximal-oriented searches (as LSM, CMG and LD) perform better for low frequency thresholds in dense datasets and for high frequency thresholds in sparse datasets. Table 7 performs a two-axes evaluation -- over generated datasets with varying densities and over a fixed dataset with varying frequency thresholds.

\begin{table}
\begin{center}
\scriptsize
\begin{tabular}{lrrr}\toprule
{\bf Search \hspace{10mm}} & Dense Gen. & Normal Gen. & Sparse Gen \\ \midrule
Simple & \textbf{--}$^\star$ & \textbf{--}$^\star$ & \textbf{--}$^\star$ \\
LSM &  3491,4 &	176,2	& 75,4 \\
CMG &   268,6 &	97,5 &	62,9 \\
LD &   1296,4 &	343,9 &	291,2 \\ \bottomrule
\end{tabular}  
\begin{tabular}{lrrrrr}
\hspace{7 mm} & \hspace{2 mm} $\theta$=0,02 &\hspace{2 mm}       $\theta$=0,05 &    \hspace{2 mm}    $\theta$=0,1 & \hspace{2 mm}       $\theta$=0,2 &     \hspace{2 mm}   $\theta$=0,4 \\ \midrule
Simple & 1020,1 &     978,9 &     792,3 &     183,0 &       6,5 \\
LSM &      5,9 &      14,1 &      26,8 &      17,0 &       1,8 \\
CMG &    4,9 &       7,4 &       9,6 &       7,6 &        1,0 \\
LD &    5,4 &       9,8 &       9,5 &       7,4 &        3,4 \\ \bottomrule
\end{tabular}
\caption{\small{Comparison of search options on generated datasets with different densities ($\theta$=0,05) and on \textsf{zoo} dataset with different frequencies using SAT4J (seconds)}}
\end{center}
\end{table}

\begin{table}
\scriptsize
\begin{center}
\begin{tabular}{llrrrrr}\toprule
           &            &       \hspace{1 mm} $\theta$=0,02 &       \hspace{1 mm} $\theta$=0,05 &        \hspace{1 mm} $\theta$=0,1 &        \hspace{1 mm} $\theta$=0,2 &        \hspace{1 mm} $\theta$=0,4 \\ \midrule
\multirow{ 2}{*}{Simple} & Number of searches &     62032 &      30369 &      9354 &       4087 &        367 \\
& Average time per search &        2,3 &        2,5 &        3,7 &        7,6 &         49 \\ \midrule
\multirow{ 2}{*}{LSM} & Number of searches &        119 &        221 &        298 &        267 &         90 \\
& Average time per search &          8 &         23 &         43 &         78 &        195 \\ \midrule
\multirow{ 4}{*}{CMG} & Number of $\alpha$-searches &        119 &        221 &        298 &        267 &         90 \\
& Average time per main search &        2,2 &        2,2 &        2,8 &        4,3 &         38 \\
& Number of $\beta$-searches &        244 &        783 &       1075 &        622 &        134 \\
& Average time per constrained search &       1,02 &       1,02 &       1,11 &       1,22 &       9,1 \\ \midrule
\multirow{ 4}{*}{LD} & Number of \textsf{unsat} searches &        32 &        32 &        32 &        32 &         32 \\
& Average time per main search &        7,2 &        8,4 &        15,3 &        27,4 &         82 \\
& Number of \textsf{sat} searches &        119 &        221 &        298 &        267 &         90 \\
& Average time per constrained search &       2,7 &       3,2 &       3,9 &       6,3 &       28,3 \\ \bottomrule
\end{tabular}  
\end{center}
\caption{\small{Search options in MiniSat+: number of searches and avg. time (miliseconds)}}
\end{table}

Simple search methods are only able to perform searches with acceptable efficiency on high frequency thresholds over very sparse datasets. In fact, even when adopting the monotonicity principle, simple searches do not scale as it is visible based on the increasing number of iterations (Table 8) as frequency thresholds decrease. Simple searches should only be adopted for an implementation that is able to promote the early selection of larger itemsets by, for instance, adjusting the polarity of the item variables. If this is the case where maximal frequent itemsets are guaranteed, simple search should be able to outperform constrained CMG.

LSM searches are only competitive when a few number of iterations is performed since each search is a lot heavier than a full $\alpha$-search (include multiple $\beta$-searches) as it has a larger space to exploit. This is usually the case where multiple similar itemsets (e.g. ACD, ACE, ADE, BCD, BDE) collapse into unique maximal frequent itemsets (e.g. ABCDE) as a result of the threshold frequency decreasing. 

CMG searches smooths the heavy computational cost of discovering multiple similar itemsets as each $\beta$-search is very light as depicted in Table 8. The average time per $\beta$-search is less than a half of an $\alpha$-search (and this relation even decreases under more larger datasets). 

Finally, LD is as competitive as CMG for medium frequency thresholds ($0.05<\theta <0.2$) on small datasets (LD performance quickly degrades with their growing size). Instead of performing $\alpha$- and $\beta$-searches, it performs one unique type of search for a fixed length of itemsets, which is a very focused search. In order to cover all possible lengths, $n$ of these searches are \textsf{unsat}. The number of \textsf{sat} searches is equal to the number of maximal frequent itemsets. A search returning \textsf{sat} has a similar performance as an $\alpha$-search, being the additional overhead added by a fixed number of \textsf{unsat} searches (understandably, not varying with the input frequency). This overhead can be critical as a search returning \textsf{unsat} is largely heavier than a search returning \textsf{sat} or a $\beta$-search (see Table 8). Therefore, the adoption of this search option essentially depends on how the number of items (defining the number of \textsf{unsat} searches in LD) compares to the number of maximal frequent itemsets (influencing the number of $\beta$-searches in CMG).

\subsection{Encoding options}

Two main observations can be derived from the experimental tests over different encodings (see Table 9). \emph{First} a clause-oriented encoding seems to be preferred over a constrained encoding. Among other aspects, the adopted constrained encoding requires the use of non-negated variables and additional clauses and variables to support later clause-removals. This is particularly true for SAT4J as this solver is not able to clausify some of these constraints. 

\begin{table}
\scriptsize
\begin{center}
\begin{tabular}{lrrrrrr}\toprule
{\bf Encoding} &       \hspace{2 mm} $\theta$=0,02 &       \hspace{2 mm} $\theta$=0,05 &        \hspace{2 mm} $\theta$=0,1 &        \hspace{2 mm} $\theta$=0,2 &        \hspace{2 mm} $\theta$=0,4 &        \hspace{2 mm} $\theta$=0,6 \\ \midrule
Clause-oriented &       4,3 &       7,9 &       9,1 &       7,1 &        1,0 &        0,3 \\
Alternative  &       4,8 &      10,9 &      29,0 &       7,0 &       1,0 &        0,2 \\
Constrained  &       5,3 &       8,8 &      46,8 &       8,4 &       2,0 &        0,4 \\
Alternative Constrained &       5,9 &      12,9 &      60,7 &       8,3 &       1,3 &        0,2 \\ \bottomrule
\end{tabular}  
\end{center}
\caption{\small{Comparison of encodings for the \textsf{zoo} dataset using SAT4J (seconds)}}
\end{table}

\begin{wrapfigure}{r}{0.6\textwidth}
  \vspace{-15pt}
  \begin{center}
    \includegraphics[width=0.6\textwidth]{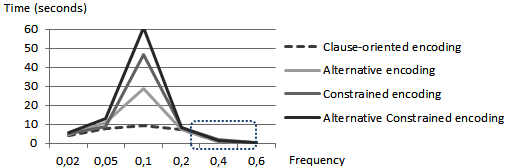}
  \vspace{-15pt}
  \end{center}
  \vspace{-15pt}
\end{wrapfigure}

\emph{Second}, alternative encodings that aim to reduce the number of clauses from $\Theta(mn)$ to $\Theta(m+n)$ (see section 3.5), may not result in significant improvements as the solver instead of having to deal with $nm$ simple binary clauses has to deal with $m$ complex constraints. The choice of whether to adopt or not this encoding mainly depends on the inputted frequency threshold (adopt for $\theta > 20$\% and avoid its use under low frequency thresholds).

\subsection{Tunning options}

As depicted in Fig.4, two simple variable polarity suggestions were undertaken. The positive suggestion is the choice for low frequency thresholds. This results from the fact that since item variables are set to true, larger itemsets tend to be initially identified. 

\begin{figure}[!htb]
\begin{center}
\includegraphics[width=7.4cm]{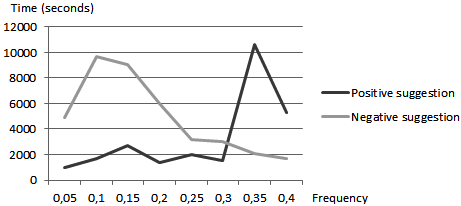}
\end{center}
\label{solution0}
\caption{\small{Positive vs. negative polarity suggestion for the \textsf{zoo.txt} dataset (miliseconds)}}
\end{figure}

However, the negative suggestion becomes the option for frequencies below 20\% in many datasets ($\sim$30\% in Fig.2). This derives from the fact that in the positive suggestion many of the initial large itemset options will not be verified with this higher thresholds, so significantly more conflicts are found within each iteration leading to an additional inefficiency related to the number of backtracks.

Advanced polarity suggestions should not only take into account an overall suggestion for all the variables, but also be able to: \emph{i)} differentiate the polarity suggestion between item variables and transaction variables, and \emph{ii)} be able to locally adapt suggestions based on task-driven heuristics (e.g. a low number of occurrences of an item variable relative to others may result in a negative polarity suggestion).

\subsection{Implementation options}

Interestingly, Fig.5 shows that each adopted SAT solver has a unique behavior when answering to the FIM problem. SAT4J is the best option when we are targeting high frequency thresholds and when mining very sparse datasets. Beyond its resolution specificities, this is also a result of accepting non-constrained encodings (including, among others, negated variables, differentiated insertion of clauses and pseudo-Boolean constraints, and clause removal). SAT4J main problems are related to memory inefficiency when dealing with large datasets and with the fact that its algorithm is more tuned to find smaller itemsets, which hampers the behavior of CMG searches since it exponentially increases the number of $\beta$-searches.

\begin{figure}[!htb]
\begin{center}
    \includegraphics[width=0.54\textwidth]{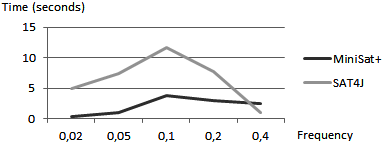}
  \end{center}
  \vspace{-5pt}
  \caption{\small{Implementation options for the \textsf{zoo} dataset under the best method}}
\end{figure}

MiniSat+ is the natural choice for the rest of the options -- dense datasets and low frequency thresholds. This is not only a consequence of the resolution methods or of C++ additional efficiency, but also a result of multiple improvements related to the solver parameterizations, with positive polarity suggestion being the most significant. 

The discussion of the behavior of other adopted solvers as PBS \cite{pbs}, BSOLO \cite{bsolo} and WBO \cite{wbo}, is out of the scope as a result of an excessive latency caused by the need to call them as executables. The successive memory refreshes among iterations and recurrent need to parse and clausify formulas distorts any potential analysis.

\subsection{Fixing phase transitions}

\begin{figure}[!htb]
\begin{subfigure}{0.52\textwidth}
\centering
    \includegraphics[width=1\textwidth]{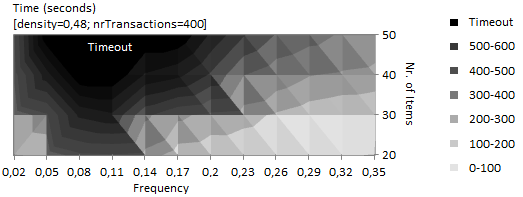}
  \caption{\scriptsize{Phase transitions for varying $\sharp$items}}
\end{subfigure}
\begin{subfigure}{0.48\textwidth}
\centering
    \includegraphics[width=1\textwidth]{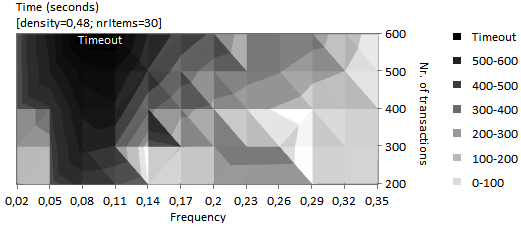}
  \caption{\scriptsize{Phase transitions for varying $\sharp$transactions}}
\end{subfigure}
\caption{\small{Phase transitions for varying size of generated datasets (using MiniSat+)}}
\end{figure}

The study of phase transitions on generated datasets with varying density (Fig.1), number of items (Fig.6a) and number of transactions (Fig.6b) led to two main observations. First, density is the property of datasets with highest impact on the performance of the proposed solution. The variation of a few percentage points can exponentially hamper the performance of our solution. Second, and understandably, the size of the dataset is also a decisive criterion affecting the behavior of our solution. A variation on the number of items is more critical than on the number of transactions. Although our solution hardly scales for a number of items above 100, for low frequencies in dense datasets or high frequencies in sparse datasets it can handle up to 10.000 transactions.


\section{Software Capabilities}

\begin{itemize}
\item flexible selection and combination of the search options, enumeration strategies, target datasets, frequencies of interest and optimization parameters (a fragment of the testing code in Java is depicted below);
\item extensible codification that gives a basis to model real-problems using simple and enumeration-centered SAT or PB:
\begin{itemize}[label=$-$]
\item \textsf{Utils} package contains a general set of encoding functionalities as the generation of \textsf{.opb} and \textsf{.cnf} files adapted to the restrictions of a particular solver; 
\item \textsf{Solver} package provides the interface to SAT solvers (supporting both a direct interface through methods invocation or via executables) and the ability to select multiple search and enumeration options in a task-independent manner; 
\end{itemize}
\item parameterizable generation of datasets and their expressive and usable adoption to test limits of performance;
\item flexible addition of pattern mining constraints (by extending the \textsf{SatPM} class);
\end{itemize}

\textsf{1: List$<$Dataset$>$ datasets = DatasetGeneration.getDatasets();}

\textsf{2: List$<$SATHandler$>$ handlers = SolverOptions.getSolvers();}

\textsf{3: List$<$Strategy$>$ strategies = StrategyOptions.getStrategies();}

\textsf{4: for(Dataset dataset : datasets)}

\textsf{5: \hspace{5mm} dataset.encodingOption(encodingID);}

\textsf{6: \hspace{5mm} for(SATHandler handler : handlers)}

\textsf{7: \hspace{10mm} handler.setPolarity(polarityID);}

\textsf{8: \hspace{10mm} for(Strategy strategy : strategies)}

\textsf{9: \hspace{15mm} for(double freq=0.01; freq$\leq$0.8; freq+=0.01)}

\textsf{10: \hspace{20mm} results.add(new StandardFIM(dataset,handler,strategy).run());}\vskip 0.4cm

\noindent The software is available in \textsf{web.ist.utl.pt/rmch/research/software}.

\end{document}